\documentclass[11pt, a4paper, logo, onecolumn,copyright]{googledeepmind}

\usepackage[authoryear, sort&compress, round]{natbib}
\title{Gemini Embedding: Generalizable Embeddings from Gemini}

\renewcommand{\today}{}

\usepackage[utf8]{inputenc} 
\usepackage[T1]{fontenc}    
\usepackage{hyperref}       
\usepackage{url}            
\usepackage{booktabs}       
\usepackage{nicefrac}       
\usepackage{microtype}      
\usepackage{amsmath}
\usepackage{graphicx}
\usepackage{multicol}
\usepackage{siunitx}
\usepackage{array}
\usepackage[nameinlink]{cleveref}
\usepackage{bbm}
\usepackage{multirow}
\usepackage{subfig}
\usepackage{soul}
\usepackage{floatrow}
\usepackage{float}
\usepackage{wrapfig}
\usepackage{blindtext}
\usepackage{tablefootnote}
\usepackage{amsfonts}
\usepackage[flushleft]{threeparttable}
\usepackage{colortbl}
\usepackage{bbding}
\usepackage{lipsum}

\usepackage{xspace}

\newfloatcommand{capbtabbox}{table}[][\FBwidth]

\newcommand{\draftonly}[1]{#1}
\newcommand{\eat}[1]{}
\renewcommand{\draftonly}[1]{}
\definecolor{darkgreen}{RGB}{0, 102, 0}

\crefformat{section}{\S#2#1#3}

\newcommand{\pairsim}{\operatorname{sim}}
\newcommand{\meanpool}{\texttt{mean\_pool}}
\newcommand{\mask}{\texttt{mask}}

\author[*]{Jinhyuk Lee}
\author[*]{Feiyang Chen}
\author[*]{Sahil Dua}
\author[*]{Daniel Cer}
\author[*]{Madhuri Shanbhogue}
\author[ \hspace{-0.2em}]{Iftekhar Naim}
\author[ \hspace{-0.2em}]{Gustavo Hern{\'{a}}ndez {\'{A}}brego}
\author[ \hspace{-0.2em}]{Zhe Li}
\author[ \hspace{-0.2em}]{Kaifeng Chen}
\author[ \hspace{-0.2em}]{Henrique Schechter Vera}
\author[ \hspace{-0.2em}]{Xiaoqi Ren}
\author[ \hspace{-0.2em}]{Shanfeng Zhang}
\author[ \hspace{-0.2em}]{Daniel Salz}
\author[ \hspace{-0.2em}]{Michael Boratko}
\author[ \hspace{-0.2em}]{Jay Han}
\author[ \hspace{-0.2em}]{Blair Chen}
\author[ \hspace{-0.2em}]{Shuo Huang}
\author[ \hspace{-0.2em}]{Vikram Rao}
\author[ \hspace{-0.2em}]{Paul Suganthan}
\author[ \hspace{-0.2em}]{Feng Han}
\author[ \hspace{-0.2em}]{Andreas Doumanoglou}
\author[ \hspace{-0.2em}]{Nithi Gupta}
\author[ \hspace{-0.2em}]{Fedor Moiseev}
\author[ \hspace{-0.2em}]{Cathy Yip}
\author[ \hspace{-0.2em}]{Aashi Jain}
\author[ \hspace{-0.2em}]{Simon Baumgartner}
\author[ \hspace{-0.2em}]{Shahrokh Shahi}
\author[ \hspace{-0.2em}]{Frank Palma Gomez}
\author[ \hspace{-0.2em}]{Sandeep Mariserla}
\author[ \hspace{-0.2em}]{Min Choi}
\author[ \hspace{-0.2em}]{Parashar Shah}
\author[ \hspace{-0.2em}]{Sonam Goenka}
\author[ \hspace{-0.2em}]{Ke Chen}
\author[ \hspace{-0.2em}]{Ye Xia}
\author[ \hspace{-0.2em}]{Koert Chen}
\author[ \hspace{-0.2em}]{Sai Meher Karthik Duddu}
\author[ \hspace{-0.2em}]{Yichang Chen}
\author[ \hspace{-0.2em}]{Trevor Walker}
\author[ \hspace{-0.2em}]{Wenlei Zhou}
\author[ \hspace{-0.2em}]{Rakesh Ghiya}
\author[ \hspace{-0.2em}]{Zach Gleicher}
\author[ \hspace{-0.2em}]{Karan Gill}
\author[ \hspace{-0.2em}]{Zhe Dong}
\author[ \hspace{-0.2em}]{Mojtaba Seyedhosseini}
\author[ \hspace{-0.2em}]{Yunhsuan Sung}
\author[ \hspace{-0.2em}]{Raphael Hoffmann}
\author[ \hspace{-0.2em}]{Tom Duerig}

\affil[ \hspace{-0.2em}]{Gemini Embedding Team, Google\footnote{See Contributions and Acknowledgments section. $^*$Equal contributions.}}

\begin{abstract}
In this report, we introduce Gemini Embedding, a state-of-the-art embedding model leveraging the power of Gemini, Google's most capable large language model. Capitalizing on Gemini's inherent multilingual and code understanding capabilities, Gemini Embedding produces highly generalizable embeddings for text spanning numerous languages and textual modalities.
The representations generated by Gemini Embedding can be precomputed and applied to a variety of downstream tasks including classification, similarity, clustering, ranking, and retrieval.
Evaluated on the Massive Multilingual Text Embedding Benchmark (MMTEB), which includes over one hundred tasks across 250+ languages, Gemini Embedding substantially outperforms prior state-of-the-art models, demonstrating considerable improvements in embedding quality. Achieving state-of-the-art performance across MMTEB's multilingual, English, and code benchmarks, our unified model demonstrates strong capabilities across a broad selection of tasks and surpasses specialized domain-specific models.
\end{abstract}

\begin{document}

\maketitle

\section{Introduction}
Embedding models, which transform inputs into dense vector representations, are pivotal for capturing semantic information across various domains and modalities.
Text embedding models represent words and sentences as vectors, strategically positioning semantically similar texts in close proximity within the embedding space~\citep{le2014distributed,reimers2019sentence,gao2021simcse}.
Recent research has focused on developing general-purpose embedding models capable of excelling in diverse downstream tasks, including information retrieval, clustering, and classification~\citep{cer2018universal,muennighoff2023mteb}.
Leveraging their vast pre-training knowledge, large language models (LLMs) have emerged as a promising avenue for constructing such general-purpose embedding models, with the potential to significantly enhance performance across a broad spectrum of applications~\citep{brown2020gpt,team2023gemini,anil2023palm}.

\begin{table}[t]
\centering
\caption{Comparison of embedding models on Massive Multilingual Embedding Benchmark: MTEB(Multilingual), MTEB(Eng, v2), and MTEB(Code). We also show results on XOR-Retrieve and XTREME-UP.
For MTEBs, we report task and type mean performances.
We report MRR@10 for XTREME-UP and Recall@5kt for XOR-Retrieve.
$^*$: Averaged over seven code tasks available for all models. $^\dagger$: For Gecko Embedding~\citep{lee2024gecko}, we evaluate text-embedding-004 on MTEB(Eng, v2), text-embedding-005 on MTEB(Code), and text-multilingual-embedding-002 on others.
}
\label{tab:main_table}
\resizebox{\columnwidth}{!}{%
\begin{tabular}{llll|llll}
\toprule
 & & {Gemini} & {Gecko} & {gte-Qwen2-} & {multilingual-e5-} & {Cohere-embed-} & {text-embedding-} \\
 & & {Embedding} & {Embedding$^\dagger$} & {7B-instruct} & {large-instruct} & {multilingual-v3.0} & {3-large} \\
\midrule
\multirow{2}{3.5cm}{\textbf{MTEB(Multilingual)} \footnotesize{\citep{enevoldsen2025mmteb}}} & Mean (Task) & \textcolor{blue}{\textbf{68.32}} & 62.13 & 62.51 & 63.23 & 61.10 & 58.92 \\
& Mean (Type) & \textcolor{blue}{\textbf{59.64}} & 54.32 & 56.00 & 55.17 & 53.31 & 51.48 \\
\cmidrule{2-8}
& - Bitext Mining & 79.32 & 70.73 & 73.92 & \textbf{80.13} & 70.50 & 62.17 \\
& - Classification & \textcolor{blue}{\textbf{71.84}} & 64.64 & 61.55 & 64.94 & 62.95 & 60.27 \\
& - Clustering & \textcolor{blue}{\textbf{54.99}} & 48.47 & 53.36 & 51.54 & 47.61 & 47.49 \\
& - Inst. Retrieval & \textcolor{blue}{\textbf{5.18}} & 4.08 & 4.94 & -0.40 & -1.89 & -2.68 \\
& - Multilabel Class. & \textcolor{blue}{\textbf{29.16}} & 22.80 & 25.48 & 22.91 & 22.74 & 22.03 \\
& - Pair Class. & 83.64 & 81.14 & \textbf{85.13} & 80.86 & 79.88 & 79.17 \\
& - Reranking & \textcolor{blue}{\textbf{65.72}} & 61.22 & 65.55 & 62.61 & 64.07 & 63.89 \\
& - Retrieval & \textcolor{blue}{\textbf{67.71}} & 59.68 & 60.08 & 57.12 & 59.16 & 59.27 \\
& - STS & \textcolor{blue}{\textbf{79.40}} & 76.11 & 73.98 & 76.81 & 74.80 & 71.68 \\
\midrule
\multirow{2}{3.5cm}{\textbf{MTEB(Eng, v2)} \footnotesize{\citep{enevoldsen2025mmteb}}} & Mean (Task) & \textcolor{blue}{\textbf{73.30}} & 69.53 & 70.72 & 65.53 & 66.01 & 66.43 \\
& Mean (Type) & \textcolor{blue}{\textbf{67.67}} & 64.82 & 65.77 & 61.21 & 61.43 & 62.15 \\
\midrule
\multirow{2}{3.5cm}{\textbf{MTEB(Code)}$^*$ \footnotesize{\citep{enevoldsen2025mmteb}}} & & \multirow{2}{*}{\textcolor{blue}{\textbf{74.66}}} & \multirow{2}{*}{65.40} & \multirow{2}{*}{56.41} & \multirow{2}{*}{57.94} & \multirow{2}{*}{51.94} & \multirow{2}{*}{58.95} \\
& & & \\
\midrule
\multirow{2}{3cm}{\textbf{XOR-Retrieve} \footnotesize{\citep{asai2021xor}}} & & \multirow{2}{*}{\textcolor{blue}{\textbf{90.42}}} & \multirow{2}{*}{65.67} & \multirow{2}{*}{N/A} & \multirow{2}{*}{N/A} & \multirow{2}{*}{N/A} & \multirow{2}{*}{68.76} \\
& & & \\
\midrule
\multirow{2}{3cm}{\textbf{XTREME-UP} \footnotesize{\citep{ruder2023xtreme}}} & & \multirow{2}{*}{\textcolor{blue}{\textbf{64.33}}} & \multirow{2}{*}{34.97} & \multirow{2}{*}{17.39} & \multirow{2}{*}{18.68} & \multirow{2}{*}{N/A} & \multirow{2}{*}{18.80} \\
& & & \\
\midrule
\textbf{Commercial Use} & & \Checkmark & \Checkmark & & & \Checkmark & \Checkmark \\
\bottomrule
\end{tabular}
}
\end{table}

The integration of LLMs has revolutionized the development of high-quality embedding models through two primary approaches.
Firstly, LLMs have been employed to refine training datasets by generating higher quality examples.
Techniques such as hard negative mining~\citep{lee2024gecko} and synthetic data generation~\citep{dai2022promptagator,wang2023improving} enable the distillation of LLM knowledge into smaller, more efficient embedding models, leading to substantial performance gains.
Secondly, recognizing that the embedding model parameters are frequently initialized from language models~\citep{Karpukhin2020DensePR,devlin2019bert}, researchers have explored leveraging LLM parameters directly for initialization~\citep{Ni2021LargeDE}.
While this approach introduces increased computational demands compared to traditional embedding models, empirical evidence suggests that utilizing strong LLMs for initialization can yield significantly superior performance~\citep{neelakantan2022text,lee2025nvembedimprovedtechniquestraining,wang2023improving}.

In this work, we introduce Gemini Embedding,\footnote{Our model is available at \url{https://ai.google.dev/gemini-api/docs/embeddings}.} a novel embedding model initialized from the powerful Gemini large language model~\citep{team2023gemini,geminiteam2024gemini15unlockingmultimodal}.
Leveraging Gemini's diverse capabilities, we train Gemini Embedding on a comprehensive suite of embedding tasks.
To construct a high-quality, heterogeneous training dataset, we employ Gemini for several critical data curation steps: filtering low-quality examples, determining relevant positive and negative passages for retrieval, and generating rich synthetic datasets.
This curated dataset facilitates training with a contrastive learning objective, enabling Gemini Embedding to learn robust semantic representations.
Building upon the success of Gecko~\citep{lee2024gecko}, we incorporate task prompts and a pre-finetuning stage to enhance performance.
Finally, we utilize Model Soup~\citep{wortsman2022model}, a simple yet effective parameter averaging technique, to combine multiple fine-tuned checkpoints, yielding a superior final embedding model.

To rigorously assess the capabilities of Gemini Embedding, we conduct extensive evaluations across a diverse spectrum of tasks and languages.
We primarily utilize the Massive Multilingual Text Embedding Benchmark (MMTEB)~\citep{enevoldsen2025mmteb}, a comprehensive test suite encompassing over 100 embedding evaluation tasks across more than 250 languages, to provide a thorough evaluation.
Notably, Gemini Embedding achieves state-of-the-art performance on MTEB(Multilingual), significantly surpassing the previous best models.
Gemini Embedding achieves a first-place ranking on the public leaderboard based on Borda rank,\footnote{\url{https://huggingface.co/spaces/mteb/leaderboard}; March 10th, 2025.} as well as on mean score averaged over tasks where it attains a score of 68.32, a substantial +5.09 improvement over the second-best model, multilingual-e5-large-instruct.
Furthermore, it achieves the highest task-type mean of 59.64, a +3.64 improvement over gte-Qwen2-7B-instruct.
As summarized in \Cref{tab:main_table}, Gemini Embedding establishes a new state-of-the-art on multiple other benchmarks such as XOR-Retrieve~\citep{asai2021xor} for cross-lingual retrieval.
Remarkably, our findings demonstrate that Gemini Embedding exhibits exceptional performance not only in high-resource languages like English but also in numerous low-resource languages, such as Macedonian.
We provide a detailed ablation study to elucidate the key factors contributing to Gemini Embedding's superior performance, offering insights into its effectiveness.
\vspace{-0.5em}
\section{Related Work}

\paragraph{Text Embedding Models}
Text embeddings are fundamental for a wide array of downstream natural language processing tasks, including semantic similarity, information retrieval, clustering, and classification.
Prior models, such as Universal Sentence Encoder~\citep{cer2018universal} and Sentence T5~\citep{ni2022sentence}, have aimed to provide general-purpose embeddings capable of handling diverse applications.
However, empirical studies have revealed limitations in their ability to generalize effectively across varied tasks and domains, highlighting the need for more robust and adaptable embedding models.
This has motivated the creation of comprehensive benchmarks like MTEB~\citep{muennighoff2023mteb,enevoldsen2025mmteb}, which emphasize novel task and domain generalization.

\vspace{-0.5em}
\paragraph{LLMs for Embedding Data Generation}
Synthetic query generation~\citep{nogueira2019document, bonifacio2022inpars, dai2022promptagator, jeronymo2023inpars} for given documents or passages has proven highly effective for creating diverse training data for embedding models.
\cite{lee2024gecko} showed that
the seed passage from which a synthetic query was generated may not be the best positive passage for that query and proposed an LLM-based approach to find better positive and negative passages.
\cite{wang2023improving} scaled up synthetic data generation over nearly one hundred languages and hundreds of thousands of tasks by prompting LLMs to first generate a diverse pool of candidate tasks and then generate data as (query, positive, hard negative) triplets conditioned on specific tasks in the pool.

\vspace{-0.5em}
\paragraph{LLMs as Embedding Models}
Pre-trained LLM encoders with bidirectional attention, such as BERT~\citep{devlin2019bert} and T5~\citep{raffel2020exploring}, have been very popular as backbones for embedding models. DPR~\citep{Karpukhin2020DensePR}, Contriever~\citep{izacard2021contriever}, Sentence-BERT~\citep{reimers2019sentence}, Language-agnostic {BERT} Sentence Embedding (LaBSE)~\citep{feng-etal-2022-language}, Sentence-T5~\citep{Ni2021LargeDE}, GTR~\citep{Ni2021LargeDE}, and E5~\citep{wang2022text} are some of the notable ones.
\cite{neelakantan2022text} initialized embedding models from decoder-only GPT-3~\citep{brown2020gpt} and adapted it for embeddings via continued contrastive pre-training. They have drastically scaled their embedding model up to 175 billion parameters, demonstrating scaling gains from pre-trained LLM backbones.

Several recent embedding models such as E5-Mistral~\citep{wang2023improving}, SFR-Mistral~\citep{meng2024sfrembedding}, BGE-ICL~\citep{li2024makingtextembeddersfewshot}, and NV-Embed~\citep{lee2025nvembedimprovedtechniquestraining} have been initialized from the Mistral-7B~\citep{jiang2023mistral7b} backbone and then further adapted as embedding models. These models generally outperform the BERT or T5 based models, showing the benefits of initializing from pre-trained LLMs.
However, their reliance on extensive in-domain training datasets has resulted in overfitting to specific benchmarks~\citep{enevoldsen2025mmteb}.
\begin{figure}[t]
\centering
\includegraphics[width=0.60\textwidth]{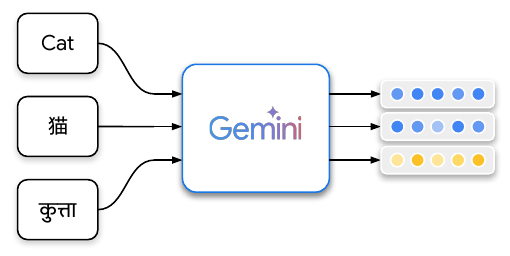}
\caption{Gemini Embedding represents text as dense vectors where semantically similar text inputs are mapped to vectors near one another in the vector space. Currently it supports more than 100+ languages, and its embeddings can be used for various tasks such as retrieval and classification.
}
\label{fig:main}
\end{figure}
\section{Gemini Embedding}
In this section we provide technical details of the Gemini Embedding model in terms of the model architecture, the objective function, and the training recipe.

\subsection{Model Architecture}
The Gemini Embedding model is built to create holistic representations of inputs for diverse downstream tasks, including retrieval, clustering, classification, and ranking by leveraging the power of Gemini.
The embedding model is initialized from Gemini and further refined.
This allows Gemini Embedding to build representations on top of the vast knowledge already present in Gemini's parameters.
In this sense, initializing the embedding model from Gemini can be seen as the "pre-training" of the Gemini Embedding model.

An input sequence $\mathbf{T}$ of $L$ tokens is processed by $\mathcal M$, a transformer with bidirectional attention initialized from Gemini, producing a sequence of token embeddings $\mathbf{T}_\mathrm{embed} = \mathcal{M}(\mathbf{T}) \in \mathbb{R}^{L \times d_\mathcal{M}}$, where $d_\mathcal{M}$ is the model dimension. To generate a single embedding representing all the information in the input, a pooler $\mathcal{P}$ is applied, $\mathbf{P}_\mathrm{embed} = \mathcal{P}(\mathbf{T}_\mathrm{embed}) \in \mathbb{R}^{d_\mathcal{M}}$.
Prior research~\citep{suganthan2025adaptingdecoder} has demonstrated that simple pooling strategies can be effective in model adaptation. Therefore we have chosen mean pooling, and simply average the token embeddings along the sequence axis. Finally, a randomly initialized linear projection $\mathit{f}$ is applied to scale the embedding to the target dimension, $\mathbf{E} = \mathit{f}(\mathbf{P}_\mathrm{embed}) \in \mathbb{R}^{d}$, where $d$ is the output embedding dimension.

\subsection{Training Objective}
\label{sec:training_objective}
The Gemini Embedding model was trained with a noise-contrastive estimation (NCE) loss with in-batch negatives. The exact loss differs slightly depending on the stage of training. In general, a training example includes a query $q_i$, a positive target $p_i^+$ and (optionally) a hard negative target $p_i^-$. Each example also has a prescribed task string $t$, for example "question answering" or "fact checking", describing the nature of the task. The query and passages are embedded as vectors in $\mathbb R^d$:
\begin{equation}
    \mathbf q_i = f(\meanpool(\mathcal M(t \oplus q_i))),\quad \mathbf p^\pm_i = f(\meanpool(\mathcal M(p^\pm_i)).
\end{equation}
Given a batch of size $B$ the loss applied to these embeddings is as follows:
\begin{equation}
    \mathcal L = \frac 1 B \sum_{i=1}^B \left[ -\log \frac{e^{\pairsim(\mathbf q_i, \mathbf p_i^+) / \tau}}{e^{\pairsim(\mathbf q_i, \mathbf p_j^-)/\tau} + \sum_{j=1}^B \mask(i,j)
    e^{\pairsim(\mathbf q_i, \mathbf p_j^+) / \tau}}\right]
\end{equation}
where $\pairsim(\mathbf x, \mathbf y)= \mathbf x^\top \mathbf y / \lVert \mathbf x \rVert \lVert \mathbf y \rVert$ is cosine similarity, and
\begin{equation}
    \mask(i,j) = \begin{cases}
    0 \quad & \text{if }q_i=q_j \text{ or } p_i^+=p_j^+,\\
    1 \quad & \text{otherwise.}
    \end{cases}
\end{equation}
This masking term is particularly relevant for classification tasks, where the number of targets (labels) is small.
The first term in the denominator is omitted if no hard negative is provided.
In contrast with Gecko \citep{lee2024gecko}, we omit the same-tower negatives \citep{moiseev2023samtone} from the loss, as we find this decreases performance for most tasks due to the potential of false negatives.

In order to support different dimensions of embeddings with a single model, we adapt the above loss using MRL \citep{kusupati2022matryoshka}, which adapts the loss above into $k$ separate losses across $k$ overlapping sub-dimensions of the embedding (e.g. multi-loss training with one loss for the first 768 embedding dimensions, another for the first 1,536 dimensions, and so on).
Gemini Embedding provides $d=3,072$ dimensional embeddings, with the MRL support on 768 and 1,536 dimensions.

\subsection{Recipe}
Initializing the embedding model from the Gemini parameters is a good starting point that leverages the language model power. This initialization can be considered a "pre-training" of the embedding model. However, in order to truly capture the generalization capabilities of initialization, we found it beneficial to leverage a two-stage training pipeline.

\paragraph{Pre-finetuning} First, the model is "pre-finetuned" on a large number of potentially noisy (query, target) pairs, omitting the hard-negative term from the loss function.
We find it beneficial to use a large batch size, as the primary objective is to adapt the parameters from autoregressive generation to encoding.
The larger batch size also provides a more stable gradient, mitigating the impact of noise in this phase of training.
Due to the larger size of the pre-finetuning dataset, pre-finetuning is performed for a substantially greater number of steps compared to fine-tuning.

\paragraph{Finetuning} Next, the model is fine-tuned on a large mixture of task-specific datasets which contain (query, target, hard negative target) triples. For this phase of training we found it beneficial to use smaller batch sizes (e.g., less than 1024), and furthermore limit each batch to a single dataset, as distinguishing a given positive target from in-batch targets from the same task provides greater signal than discerning (say) a retrieval target from a classification label.
We perform a grid search of various training hyperparameters, including the inclusion and exclusion of components of the mixture, to obtain candidate checkpoints.

\paragraph{Model Soup} To obtain additional generalization performance, we averaged the parameters obtained from individual fine-tuning runs. We experimented with different combinations of parameters, including averaging checkpoints from the same training run \citep{izmailov2018averaging}, from different training runs \citep{wortsman2022model}, as well as various weighted averages. The final set of ingredient checkpoints were obtained through a combination of intentional data variation as well as manual checkpoint selection and experimentation.

\section{Datasets}
\label{sec:Datasets}
Our training data mixture contains diverse multilingual embedding tasks as well as code retrieval tasks.
Gemini is used in three different ways to improve the quality of our data: synthetic data generation, data filtering, and hard negative mining.

\vspace{-0.5em}
\subsection{Training Data Mixture}
\paragraph{Pre-finetuning}
Our pre-finetuning stage aims to maximize the exposure of diverse training datasets to Gemini Embedding models.
We leverage a billion-scale web corpus and used title and passage pairs as input and positive target pairs, similar to some prior work~\citep{neelakantan2022text,lee2024gecko}.
Despite being very simple, this technique is consistently found to be effective even when the embedding model is initialized from an LLM.

\vspace{-0.5em}
\paragraph{Fine-tuning}
For fine-tuning, we prepare three different mixtures aiming for task diversity, language diversity, and coding capability.
For task diversity, we use a subset of academic datasets used by Gecko~\citep{lee2024gecko} as well as several synthetic datasets introduced in \Cref{section:gemini_data}.
Unlike existing models on the classic MTEB~\citep{muennighoff2023mteb}, we excluded many in-domain MTEB datasets, which improved the performance only on their own test split mainly due to train-test leakage or dataset bias.
The training mixture rate was decided based on a fine-grained grid search, initialized from the optimal number of training steps to converge on each training dataset.

\vspace{-0.5em}
\subsection{Improving Data Quality with Gemini}\label{section:gemini_data}
\paragraph{Synthetic Data Generation}
Recent embedding evaluation benchmarks such as MMTEB~\citep{enevoldsen2025mmteb} contain many different tasks other than retrieval.
We diversify and improve our training mixture by adding synthetically generated datasets for two task types: retrieval and classification.
For retrieval, we extended our prior work on synthetic data generation using Gemini enhanced adaptations of FRet~\citep{lee2024gecko} and SWIM-IR~\citep{thakur-etal-2024-leveraging}. Using few-shot prompting, we first use Gemini to generate synthetic queries for web passages followed by a Gemini auto-rater to filter lower-quality examples (e.g., unrealistic search queries).
For classification, we generate synthetic counterfactual, sentiment, and review classification datasets in English.
To increase the quality of these synthetic datasets we developed multi-stage prompting strategies, such as conditioning on synthetic user, product, or movie generations in a hierarchical manner and sampling from the tail of longer lists of generations, as diversity naturally increases with generation length.

\vspace{-0.5em}
\paragraph{Data Filtering}
Our training data mixture includes many human-annotated datasets.
We noticed that many retrieval datasets have quality issues of incorrect positive or negative targets for a query.
We use Gemini to filter such bad examples.
Based on our few-shot prompting for data quality assessment, we remove low quality examples.

\vspace{-0.5em}
\paragraph{Hard Negative Mining}
A standard technique when training embedding models is to mine "hard negatives," i.e. targets which are semantically similar to a true positive target but do not answer the query~\citep{reddi2019stochastic}.
We mine hard negatives for our retrieval datasets using Gemini.
We first train a Gemini-initialized embedding model without using any hard negatives.
Based on this initial embedding model, we retrieve top $k$ nearest neighbors for each query.
Each nearest neighbor is then scored by Gemini along with the query.
We follow \citet{lee2024gecko} and employ two different prompting strategies---graded classification and query likelihood---combining the scores with Reciprocal Rank
Fusion (RRF)~\citep{cormack2009reciprocal}.
We found that the lowest-scoring nearest neighbors, (the $k$-th neighbor after being sorted by Gemini scores) serve as the best hard negatives.

\begin{table}[t]
\centering
\caption{Performance of top leaderboard models on MTEB(Multilingual).}
\label{tab:leaderboard_multilingual}
\resizebox{\columnwidth}{!}{%
\begin{tabular}{lccc|ccccccccc}
\toprule
& & \multirow{2}{*}{\parbox{1cm}{\centering{\textbf{Mean}\\\textbf{(Task)}}}} & \multirow{2}{*}{\parbox{1cm}{\centering \textbf{Mean}\\\textbf{(Type)}}} & \multirow{2}{*}{\parbox{1.2cm}{\centering Bitext\\Mining}} & & & \multirow{2}{*}{\parbox{1.2cm}{\centering Inst.\\Retrieval}} & \multirow{2}{*}{\parbox{1.2cm}{\centering Multi.\\Class.}} & \multirow{2}{*}{\parbox{1.2cm}{\centering Pair.\\Class.}} \\
Model Name & \textbf{Rank} &  &  & & {Class.} & {Clus.} & & & & {Rerank.} & {Retrieval} & {STS} \\
\midrule
{Gemini Embedding} & \textbf{1} & \textcolor{blue}{\textbf{68.3}} & \textcolor{blue}{\textbf{59.6}} & 79.3 & \textcolor{blue}{\textbf{71.8}} & \textcolor{blue}{\textbf{55.0}} & \textcolor{blue}{\textbf{5.2}} & \textcolor{blue}{\textbf{29.2}} & 83.6 & \textcolor{blue}{\textbf{65.6}} & \textcolor{blue}{\textbf{67.7}} & \textcolor{blue}{\textbf{79.4}} \\
\midrule

Linq-Embed-Mistral & 2 & 61.5 & 54.2 & 70.3 & 62.2 & 51.3 & 0.9 & 24.8 & 80.4 & 64.4 & 58.7 & 74.9 \\
gte-Qwen2-7B-instruct & 3 & 62.5 & 56.0 & 73.9 & 61.6 & 53.4 & 4.9 & 25.5 & \textbf{85.1} & 65.6 & 60.1 & 74.0 \\
multilingual-e5-large-instruct & 4 & 63.2 & 55.2 & \textbf{80.1} & 64.9 & 51.5 & -0.4 & 22.9 & 80.9 & 62.6 & 57.1 & 76.8 \\
SFR-Embedding-Mistral & 5 & 60.9 & 54.0 & 70.0 & 60.0 & 52.6 & 0.2 & 24.6 & 80.3 & 64.2 & 59.4 & 74.8 \\
GritLM-7B & 6 & 60.9 & 53.8 & 70.5 & 61.8 & 50.5 & 3.5 & 22.8 & 79.9 & 63.8 & 58.3 & 73.3 \\
text-multilingual-embedding-002 & 7 & 62.1 & 54.3 & 70.7 & 64.6 & 48.5 & 4.1 & 22.8 & 81.1 & 61.2 & 59.7 & 76.1 \\
GritLM-8x7B & 8 & 60.5 & 53.4 & 68.2 & 61.6 & 50.9 & 2.4 & 24.4 & 79.7 & 62.6 & 57.5 & 73.2 \\
e5-mistral-7b-instruct & 9 & 60.3 & 53.2 & 70.6 & 60.3 & 51.4 & -0.6 & 22.2 & 81.1 & 63.8 & 55.8 & 74.0 \\
Cohere-embed-multilingual-v3.0 & 10 & 61.1 & 53.3 & 70.5 & 63.0 & 47.6 & -1.9 & 22.7 & 79.9 & 64.1 & 59.2 & 74.8 \\
gte-Qwen2-1.5B-instruct & 11 & 59.5 & 52.8 & 62.5 & 58.3 & 52.6 & 0.7 & 24.0 & 81.6 & 62.6 & 60.8 & 71.6 \\
bilingual-embedding-large & 12 & 60.9 & 53.0 & 73.6 & 62.8 & 47.2 & -3.0 & 22.4 & 79.8 & 61.4 & 55.1 & 77.8 \\
\bottomrule
\end{tabular}
}
\end{table}
\section{Evaluation}
Gemini Embedding is assessed on a comprehensive collection of task types, domains, languages, and language pairs (e.g., Hindi queries retrieving English content) using benchmark evaluations from the Massive Multilingual Text Embedding Benchmark, MMTEB~\citep{enevoldsen2025mmteb}, and the cross-lingual benchmarks XTREME-UP~\citep{ruder2023xtreme} and XOR-Retrieve~\citep{asai2021xor}.

\subsection{Benchmarks and Tasks}
MMTEB consists of a large collection of individual evaluation tasks covering 250+ languages and 10 task types: Bitext Mining, Classification, Clustering, Instruction Retrieval, Multilabel Classification, Pair Classification, Reranking, Retrieval, STS, and Summarization. 
Our MMTEB evaluations include 164 individual evaluation tasks consisting of 132 evaluation tasks for MTEB(Multilingual), 41 tasks for MTEB(Eng, v2), and 12 code retrieval tasks for MTEB(Code).
Notably, MTEB(Multilingual) contains 250+ languages.
XOR-Retrieve and XTREME-UP provide cross-lingual retrieval evaluations, with XOR-Retrieve pairing English passages with retrieval queries in 7 different languages and XTREME-UP similarly pairing English passages with queries in 20 underrepresented Indo-European languages.

\subsection{Overall Performance}
Gemini Embedding's overall performance along with that of other top performing models is presented in \Cref{tab:main_table} on the following evaluations: three benchmarks from MMTEB, MTEB(Multilingual), MTEB(Eng, v2), MTEB(Code); and the two cross-lingual benchmarks XOR-Retrieve and XTREME-UP.

Gemini Embedding establishes a new state-of-the-art in performance, achieving the highest overall performance on the MTEB(Multilingual) leaderboard (March 10th, 2025) with a substantial performance lead over all previous top performing models on each of the overall metrics summarizing aggregate performance across tasks: Task Mean (equal weighting of all tasks): 68.32, Task Type Mean (equal weighting of all task types): 59.64, and Borda rank \#1 (official leaderboard ranking metric). Gemini Embedding's performance advantage is not limited to just MTEB(Multilingual). Within a single unified model and shared embedding space, Gemini Embedding's capabilities allow it to achieve: (i) \textbf{\#1 ranking on MTEB(Multilingual)}, (ii) \textbf{\#1 ranking on MTEB(Eng, v2)}, (iii) \textbf{\#1 ranking on MTEB(Code)}, and (iv) \textbf{excellent cross-lingual retrieval on XOR-Retrieve and XTERME-UP}, advancing the state-of-the-art for general-purpose embeddings as cross-lingual representations.

\begin{table}[t]
\centering
\caption{Performance of top leaderboard models on MTEB(Eng, v2).
}
\label{tab:leaderboard_eng}
\resizebox{\columnwidth}{!}{%
\begin{tabular}{lccc|ccccccc}
\toprule

& & \multirow{2}{*}{\parbox{1cm}{\centering{\textbf{Mean}\\\textbf{(Task)}}}} & \multirow{2}{*}{\parbox{1cm}{\centering \textbf{Mean}\\\textbf{(Type)}}} & & &
\multirow{2}{*}{\parbox{1.2cm}{\centering Pair.\\Class.}} & & & &
\\

Model Name & \textbf{Rank} & & & {Class.} & {Clus.} & & {Rerank.} & {Retrieval} & {STS} & {Summ.}\\
\midrule
        {Gemini Embedding} & \textbf 1 & \textcolor{blue}{\textbf{73.3}} & \textcolor{blue}{\textbf{67.7}} & 90.1 & 59.4 & 87.7 & 48.6 & \textcolor{blue}{\textbf{64.4}} & \textcolor{blue}{\textbf{85.3}} & \textcolor{blue}{\textbf{38.3}} \\
        \midrule
        Linq-Embed-Mistral & 2 & 69.8 & 65.3 & 83.0 & 54.1 & 88.4 & 49.4 & 60.1 & 84.7 & 37.3 \\
        jasper\_en\_vision\_language\_v1 & 3 & 71.4 & 66.7 & \textbf{90.3} & \textbf{60.5} & 88.1 & 50.0 & 56.1 & 84.4 & 37.2 \\
        SFR-Embedding-Mistral & 4 & 69.3 & 64.9 & 80.5 & 54.9 & 88.6 & 50.2 & 59.3 & 84.8 & 36.3 \\
        NV-Embed-v2 & 5 & 69.8 & 65.0 & 87.2 & 47.7 & \textbf{88.7} & 49.6 & 62.8 & 83.8 & 35.2 \\
        text-embedding-005 (Gecko) & 6 & 69.6 & 64.8 & 86.0 & 51.9 & 87.6 & 48.8 & 58.8 & 85.2 & 35.1 \\
        text-embedding-004 (Gecko) & 7 & 69.5 & 64.8 & 86.0 & 51.5 & 87.7 & 48.5 & 59.1 & 84.8 & 36.1\\
        gte-Qwen2-7B-instruct & 8 & 70.7 & 65.8 & 88.5 & 59.0 & 85.9 & \textbf{50.5} & 58.1 & 82.7 & 35.7 \\
        e5-mistral-7b-instruct & 9 & 68.0 & 64.0 & 79.9 & 51.4 & 88.4 & 49.8 & 57.6 & 84.3 & 36.6 \\
        stella\_en\_400M\_v5 & 10 & 69.4 & 64.8 & 88.3 & 57.7 & 87.2 & 49.6 & 52.7 & 83.9 & 34.5 \\
        stella\_en\_1.5B\_v5 & 11 & 69.4 & 65.3 & 89.4 & 57.1 & 88.0 & 50.2 & 52.4 & 83.3 & 36.9 \\
        gte-Qwen2-1.5B-instruct & 12 & 67.2 & 63.3 & 85.8 & 53.5 & 87.5 & 49.3 & 50.3 & 82.5 & 33.9 \\
\bottomrule
\end{tabular}
}\vspace{-0.5em}
\end{table}
\begin{table}[t]
\centering
\caption{Performance of top leaderboard models on MTEB(Code).
} 
\label{tab:leaderboard_code}
\resizebox{\columnwidth}{!}{%
\begin{tabular}{lccc|cccccccc}
\toprule

& & \multirow{2}{*}{\parbox{1cm}{\centering{\textbf{Mean}\\\textbf{All}}}} & \multirow{2}{*}{\parbox{1cm}{\centering{\textbf{Mean}\\\textbf{\mbox{-COIR}}}}} & & & & & & & & 
\\

Model Name & \textbf{Rank} & & & {AppsR.} & {COIR} & {CESR} & {CSNCCR} & {CSNR} & {CTOC} & {CTODL} & {CQA} \\
\midrule
        {Gemini Embedding} & \textbf 1 & \textcolor{blue}{\textbf{75.5}} & \textcolor{blue}{\textbf{74.7}} & \textcolor{blue}{\textbf{93.8}}	& 81.1 & \textcolor{blue}{\textbf{81.6}} & 84.7 & 91.3 & 89.5 & 31.5 & 50.2 \\
\midrule
        inf-retriever-v1-1.5b & 2 & 62.9 & 60.6 & 38.9 & 78.6 & 67.2 & 75.5 & 90.9 & 85.0 & 33.8 & 33.1 \\
        text-embedding-005 (Gecko) & 3 & 63.3 & 65.4 & 91.3 & 48.4 & 54.4 & 55.7 & 87.2 & 82.8 & 34.4 & \textbf{52.2} \\
        voyage-code-3 & 4 & - & - & 93.6 & \textbf{89.4} & - & \textbf{90.1} & \textbf{94.0} & \textbf{95.0} & \textbf{38.6} & 34.5 \\
        NV-Embed-v2 & 5 & - & 59.4 & 29.1 & -  & 74.0 & 68.8 & 86.6 & 89.1 & 33.4 & 34.8 \\
        voyage-3 & 6 & - & 67.3 & 73.0 & - & 75.6 & 77.9 & 92.3 & 89.9 & 33.9 & 28.7 \\
        GritLM-7B & 7 & - & 62.4 & 35.1 & -  & 74.6 & 86.7 & 86.7 & 89.2 & 33.0 & 31.2 \\
        KaLM-emb.-mling.-mini-v1 & 8 & - & 57.4 & 46.8 & -  & 60.0 & 59.5 & 88.0 & 79.9 & 34.0 & 33.6 \\
        text-embedding-3-large & 9 & - & 59.0 & 28.4 & - & 71.1 & 73.2 & 90.5 & 84.3 & 34.2 & 31.0 \\
        NV-Embed-v1 & 10 & - & 57.7 & 30.3 & - & 70.8 & 65.1 & 85.8 & 85.1 & 33.1 & 33.4 \\
        SFR-Embedding-Mistral & 11 & - & 56.7 & 26.1 & - & 68.8 & 64.5 & 86.7 & 83.5 & 32.9 & 34.3 \\
        Linq-Embed-Mistral & 12 & - & 57.5 & 30.2 & - & 70.6 & 64.5 & 87.1 & 84.9 & 32.8 & 32.6 \\
\bottomrule
\end{tabular}
}
\end{table}

\paragraph{MTEB(Multilingual) leaderboard}
In \Cref{tab:leaderboard_multilingual}, Gemini Embedding is compared with top-ranked models from MTEB(Multilingual).
Achieving the highest Borda rank and excellent overall performance across task types, Gemini Embedding particularly excels at Classification (+9.6), Clustering (+3.7) and Retrieval (+9.0)  compared to the second-best model.

\vspace{-0.5em}
\paragraph{MTEB(Eng, v2) leaderboard} Comparing with top-ranked MTEB(Eng, v2) leaderboard models in \Cref{tab:leaderboard_eng}, Gemini Embedding achieves the highest Borda rank and great overall performance across task types, with particularly striking performance improvements on Classification (+7.1), Clustering (+5.3), and Retrieval (+4.3)  compared to the second-best model.

\begin{table}[t]
\centering
\caption{Performance of top multilingual models on XTREME-UP (MRR@10).}
\resizebox{\columnwidth}{!}{%
\begin{tabular}{ll|c*{20}{c}}
\toprule
 
 & Average & as & bho & brx & gbm & gom & gu & hi & hne & kn & mai & ml & mni & mr & mwr & or & pa & ps & sa & ta & ur\\
\midrule
{{Gemini Embedding}} & \textcolor{blue}{\textbf{64.3}} & \textcolor{blue}{\textbf{69.2}} & \textcolor{blue}{\textbf{66.4}} & \textcolor{blue}{\textbf{25.7}} & \textcolor{blue}{\textbf{64.9}} & \textcolor{blue}{\textbf{65.5}} & \textcolor{blue}{\textbf{70.3}} & \textcolor{blue}{\textbf{69.1}} & \textcolor{blue}{\textbf{68.3}} & \textcolor{blue}{\textbf{69.5}} & \textcolor{blue}{\textbf{68.4}} & \textcolor{blue}{\textbf{70.8}} & \textcolor{blue}{\textbf{44.4}} & \textcolor{blue}{\textbf{68.8}} & \textcolor{blue}{\textbf{66.5}} & \textcolor{blue}{\textbf{65.8}} & \textcolor{blue}{\textbf{69.5}} & \textcolor{blue}{\textbf{61.9}} & \textcolor{blue}{\textbf{68.1}} & \textcolor{blue}{\textbf{68.6}} & \textcolor{blue}{\textbf{64.8}} \\
{{Gecko i18n Embedding}} & 35.0 & 31.9 & 39.7 & 3.8 & 37.4 & 26.0 & 42.9 & 46.3 & 42.0 & 41.6 & 44.1 & 45.5 & 9.4 & 41.5 & 40.7 & 19.4 & 40.9 & 33.0 & 35.9 & 40.5 & 37.0 \\
\midrule
voyage-3-large & 39.2 & 34.3 & 44.8 & 7.9 & 46.6 & 27.1 & 46.7 & 54.3 & 45.3 & 41.5 & 48.3 & 45.3 & 19.2 & 45.5 & 47.9 & 32.3 & 48.4 & 26.8 & 40.0 & 36.0 & 45.6 \\
Linq-Embed-Mistral & 24.6 & 23.8 & 38.1 & 8.6 & 37.0 & 21.7 & 11.6 & 44.2 & 39.7 & 21.7 & 38.5 & 10.2 & 14.7 & 31.4 & 36.2 & 10.7 & 8.3 & 13.8 & 37.7 & 14.3 & 29.3 \\
multiling.-e5-large-instr. & 18.7 & 21.2 & 21.9 & 1.5 & 19.3 & 8.7 & 13.9 & 30.6 & 22.6 & 24.2 & 24.0 & 8.6 & 6.3 & 23.0 & 19.8 & 17.3 & 24.5 & 15.9 & 19.1 & 22.9 & 28.2 \\
gte-Qwen2-7B-instruct & 17.4 & 14.7 & 22.7 & 5.4 & 23.0 & 7.0 & 19.1 & 30.4 & 19.1 & 16.2 & 25.9 & 21.7 & 7.2 & 23.8 & 24.0 & 11.3 & 19.2 & 11.0 & 21.1 & 9.7 & 15.5 \\
text-embedding-3-large & 18.8 & 18.2 & 28.8 & 3.3 & 28.4 & 11.1 & 14.6 & 40.4 & 29.3 & 17.1 & 31.1 & 15.6 & 2.9 & 25.5 & 28.7 & 8.3 & 11.3 & 6.8 & 26.6 & 6.0 & 22.0 \\
\bottomrule
\end{tabular}
}
\label{tab:xtreme_up_results}
\end{table}
\begin{figure}[t]
\centering
\includegraphics[width=0.95\textwidth]{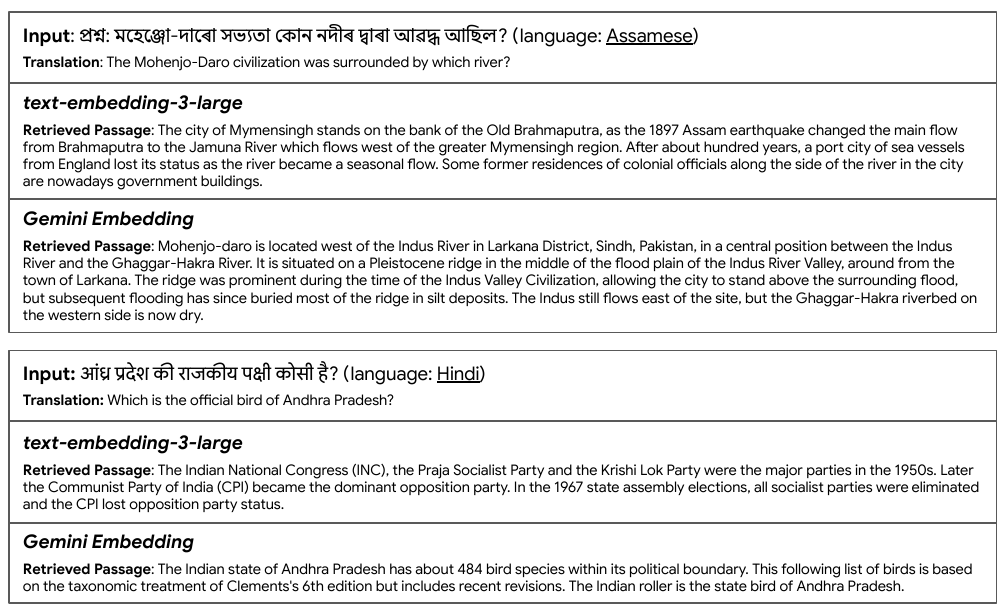}
\caption{Gemini Embedding supports cross-lingual retrieval where different languages can be used for queries and passages.
We show two examples from XTREME-UP showing the strong cross-lingual retrieval capability of Gemini Embedding. Despite Assamese being a relatively low-resource language and the Hindi query having a typo, the Gemini Embedding model correctly understood the key entities and the contexts in the queries and retrieved the correct passages.
}
\label{fig:qual_examples}
\end{figure}

\vspace{-0.5em}
\paragraph{MTEB(Code) leaderboard} The eight tasks present on the MTEB(code) leaderboard, which excludes the four additional MTEB(code) tasks CodeFeedbackMT, CodeFeedbackST, StackOverflowQA, and SyntheticText2SQL, are shown in \Cref{tab:leaderboard_code}.
Only a few models, including both Gemini Embedding and Google's Gecko model, have been submitted to the MTEB(Code) leaderboard with evaluations over all tasks.
On the MTEB(Code) leaderboard, Gemini Embedding once again achieves the highest Borda rank and mean performance across all eight evaluation tasks.
Since the majority of other top models on MTEB(Code) are missing COIRCodeSearchNetRetrieval (COIR), we also report the mean performance over the seven remaining tasks, \textbf{Mean -COIR}.
Gemini Embedding still achieves the best mean performance over the seven \textbf{Mean -COIR} evaluation tasks.

\paragraph{XTREME UP} The performance of Gemini Embedding along with the top-performing multilingual models on XTREME-UP cross-lingual retrieval is presented in \Cref{tab:xtreme_up_results}. XTREME-UP requires mapping queries in 20 underrepresented languages to English passages. Gemini Embedding demonstrates a remarkable improvement in cross-lingual retrieval with its general-purpose embeddings.

\subsection{Qualitative Examples}
In \Cref{fig:qual_examples}, we show examples from XTREME-UP that show the cross-lingual retrieval capability of Gemini Embedding.
The two queries are given in Assamese and Hindi, and the task is to retrieve relevant English passages that contain the answers.
Each query without any translation is encoded and the highest-scoring English passages are retrieved using cosine similarity.
Gemini Embedding found the right passages showcasing its strong capability on multilingual and cross-lingual tasks.

\section{Ablation Study}
To better understand how Gemini Embedding achieves great performance across many different tasks and languages, we provide a systematic analysis of our training recipe.

\begin{table}[t]
\centering
\caption{Results using different training mixtures for MTEBs (task mean), XTREME-UP (MRR@10), and XOR-Retrieve (Recall@5kt). Using a Gemini foundation, the English Only mixture is able to achieve good performance on MTEB(Multilingual), MTEB(Eng, v2) and XOR-Retrieve. Multilingual fine-tuning helps the most on the long-tail languages in XTREME-UP. Ablations exclude model souping.
}
\label{tab:main_ablation}
\resizebox{1.00\columnwidth}{!}{%
\begin{tabular}{lcccccc}
\toprule
& {MTEB(Multilingual)} & {MTEB(Eng, v2)} & {MTEB(Code)} & {XOR-Retrieve} & {XTREME-UP} \\
\midrule
Gemini Embedding & \textcolor{blue}{\textbf{68.32}} & \textcolor{blue}{\textbf{73.28}} & \textcolor{blue}{\textbf{74.66}} & \textcolor{blue}{\textbf{90.42}} & 64.33 \\
\midrule
\textit{Pre-Finetuning} \\
\midrule
Pre-finetuning Only & \textbf{48.89} & \textbf{50.99} & \textbf{46.18} & 76.64 & 21.22 \\
No Training & 30.55	& 28.17 & 9.86 & - & - \\
\midrule
\textit{Fine-tuning Mixtures} \\
\midrule
English Only (Diverse Task) & \textbf{66.75} & \textbf{72.77} & 58.68 & 85.70 & 49.34 \\
Multilingual Only (Retrieval) & 58.24 & 61.88 & 58.75 & \textbf{89.00} & \textbf{65.06} \\
Code Only (Retrieval) & 60.20 & 62.25 & \textbf{72.08} & 82.16 & 34.74 \\
\bottomrule
\end{tabular}
}
\end{table}
\begin{table}[t]
\centering
\caption{Results on MTEB classification using synthetic datasets. Self-training on Gemini generated training data dramatically improves model performance, \textbf{+17.6}. Ablation models exclude souping. {\small \mbox{$^*$ Gecko} training mixtures include training sets provided by several classification tasks from Huggingface.}}
\label{tab:synthetic_data}
\resizebox{\columnwidth}{!}{%
\begin{tabular}{ll|cccc}
\toprule
& \textbf{Average} & {AmazonCounterfactual} & {AmazonPolarity} & {AmazonReviews} & {Emotion} \\
\midrule
\textbf{w/o Synthetic} & 57.57 & 65.43 & 67.29 & 48.84 & 48.70 \\
\textbf{w/ Synthetic} & \textbf{75.17} (+17.6) & \textbf{91.30} & \textbf{96.51} & \textbf{57.00} & \textbf{55.90} \\
\midrule
{Gecko Embedding} & 66.78 & 66.52 & \textbf{97.28}$^*$ & 51.24 & 52.09 \\
{Gemini Embedding} & \textcolor{blue}{\textbf{76.09}} & \textcolor{blue}{\textbf{92.70}} & 96.10 & \textcolor{blue}{\textbf{59.30}} & \textcolor{blue}{\textbf{56.27}} \\
\bottomrule
\end{tabular}
}
\end{table}

\subsection{Does Gemini Embedding Generalize to Multilingual Tasks?}
In \Cref{tab:main_ablation}, we show how Gemini Embedding can generalize over different languages and tasks.
In the middle rows, we show our model's performance before fine-tuning: no training and pre-finetuning only.
Pre-finetuning greatly improves the performance across multiple benchmarks.
The bottom rows show the effect of further fine-tuning the pre-finetuned checkpoints.
We find that training on the English-only mixture still achieves very strong performance on MTEB(Multilingual) where the evaluations are mostly zero-shot. 
Remarkably, even when training our model on the English-only mixture, we are able to outperform the top embedding models on XTREME-UP.\footnote{+10.1 MMR@10 for English-only fine-tuning in \Cref{tab:main_ablation} vs. the top performing non-Gemini model in \Cref{tab:xtreme_up_results}}
This shows Gemini Embedding can generalize over different languages even if its training mixture contains only a single language.
On the other hand, our multilingual-only mixture consists of only retrieval datasets but not other task types such as classification.
Its lower score indicates that task diversity matters more than language diversity for fine-tuning in Gemini Embedding.

\begin{table}[t]
\centering
\caption{Results on filtering the MIRACL datasets. We show that proper filtering of retrieval datasets using LLMs can greatly improve the performance.}
\resizebox{\columnwidth}{!}{%
\begin{tabular}{ll|c*{17}{c}}
\toprule
 & \textbf{Average} & ar & bn & de & en & es & fa & fi & fr & hi & id & ja & ko & ru & sw & te & th & yo & zh \\
\midrule
\textbf{w/o Filtering} & 59.8 & \textbf{74.8} & 71.5 & 46.5 & 54.6 & 44.5 & 51.3 & 66.8 & 46.0 & \textbf{59.4} & 34.9 & 56.9 & 55.1 & 62.6 & 69.7 & 73.6 & 71.9 & \textbf{85.0} & 51.7 \\
\textbf{w/ Filtering} & \textbf{63.7} (+3.9) & 74.2 & \textbf{74.7} & \textbf{52.9} & \textbf{54.7} & \textbf{47.3} & \textbf{55.5} & \textbf{74.7} & \textbf{49.5} & 59.3 & \textbf{47.1} & \textbf{61.9} & \textbf{63.3} & \textbf{64.8} & \textbf{76.0} & \textbf{75.0} & \textbf{75.0} & 83.3 & \textbf{57.0} \\
\midrule
{{Gecko Embedding}} & 56.2 & 64.3 & 66.7 & 49.1 & 45.3 & 48.5 & 49.2 & 65.2 & 45.1 & 55.0 & 44.7 & 52.6 & 57.5 & 55.1 & 67.4 & 74.5 & 66.5 & 54.0 & 50.7 \\
{{Gemini Embedding}} & \textcolor{blue}{\textbf{70.1}} & \textcolor{blue}{\textbf{78.3}} & \textcolor{blue}{\textbf{79.0}} & \textcolor{blue}{\textbf{59.8}} & \textcolor{blue}{\textbf{58.7}} & \textcolor{blue}{\textbf{57.0}} & \textcolor{blue}{\textbf{60.9}} & \textcolor{blue}{\textbf{78.0}} & \textcolor{blue}{\textbf{55.6}} & \textcolor{blue}{\textbf{65.4}} & \textcolor{blue}{\textbf{54.3}} & \textcolor{blue}{\textbf{75.1}} & \textcolor{blue}{\textbf{68.9}} & \textcolor{blue}{\textbf{73.4}} & \textcolor{blue}{\textbf{81.0}} & \textcolor{blue}{\textbf{80.5}} & \textcolor{blue}{\textbf{80.8}} & \textcolor{blue}{\textbf{88.8}} & \textcolor{blue}{\textbf{65.7}} \\
\bottomrule
\end{tabular}
}
\label{tab:miracl_results}
\end{table}
\begin{figure}[t]
\centering
\includegraphics[width=0.85\textwidth]{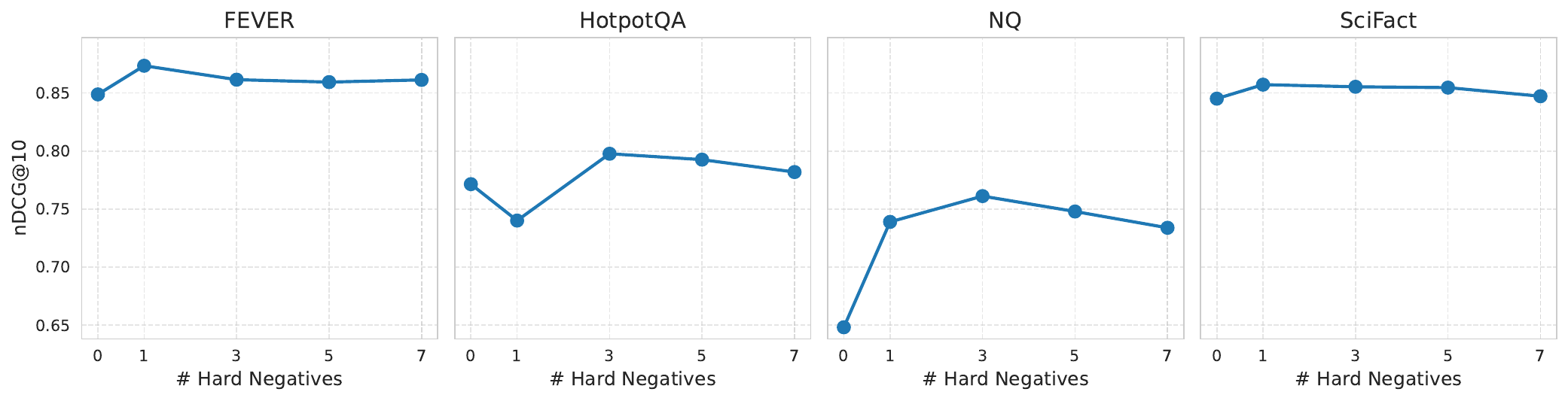}
\caption{Results on retrieval datasets with different number of hard negatives. We show that our hard negatives are mostly useful.}
\label{fig:hn_mining}
\end{figure}

\subsection{How Does Gemini Improve Data Quality?}
\vspace{-0.3em}
\paragraph{Synthetic Data Generation}
We show the effectiveness of our multi-stage prompting strategy to create diverse, realistic synthetic classification datasets in \Cref{tab:synthetic_data}.
Note that these are zero-shot synthetic datasets, so no actual examples from the original datasets were used when prompting Gemini.
Training on our synthetic classification datasets greatly improves the performance on all datasets.
We find that the performance with synthetic datasets can match the performance of in-domain datasets (e.g. Gecko on AmazonPolarity), and our multi-stage prompting strategy even allows for controllable generation, raising the possibility of reducing bias compared to real data.

\vspace{-1.0em}
\paragraph{Data Filtering}
We use Gemini to filter retrieval datasets.
We test filtering the MIRACL~\citep{zhang2023miracl} training datasets, which contain retrieval datasets in 18 different languages, and measure the impact of training on the filtered dataset.
\Cref{tab:miracl_results} shows that filtered results consistently show better results across different languages showing only minor drops for some languages.
As demonstrated in \Cref{tab:main_ablation}, our English mixture helps to improve the quality on multilingual tasks, making Gemini Embedding the best in \Cref{tab:miracl_results} as well.

\vspace{-1.0em}
\paragraph{Hard Negative Mining}
We examine the quality of our hard negatives selected by Gemini.
As demonstrated in \Cref{fig:hn_mining}, incorporating hard negatives generally enhances our model's retrieval performance across the four datasets.
However, excessive hard negatives often led to overfitting, causing performance degradation for retrieval tasks.
Future work will explore regularization techniques and better hard negative sampling strategies to address overfitting.

\vspace{-1.0em}
\section{Future Work}
Beyond the text embedding capabilities described here, we will explore extending the embedding capabilities for other modalities like image, video, and audio. We want to leverage the powerful multi-modal capabilities of Gemini to make the Gemini Embedding model comprehensive~\citep{jiang2024vlm2vec} in terms of representing different combinations of modalities together in a single embedding space. This will require curating multi-modal data tasks suitable for learning generalizable representations. We will also explore training recipes that will balance the performance of a single model across different uni-modal and multi-modal capabilities.

\section{Conclusion}

Gemini Embedding is a unified, general-purpose, and highly-capable embedding model that capitalizes on the strong capabilities of Gemini to advance the state-of-the-art in representation learning. Building on an excellent foundation provided by Gemini's multilingual and code understanding capabilities, Gemini Embedding generates a versatile encoding of model inputs into representations with a wide range of capabilities over many languages, domains, and task types including: classification, similarity search, clustering, ranking, and retrieval. Gemini Embedding both adapts the capabilities of Gemini to representation learning and uses Gemini itself to generate many of the training sets for this adaptation. The resulting representations benefit from the underlying capabilities of Gemini itself while also being efficient to precompute, cache, and re-use them. Efficiently cacheable and reusable representations unlock the ability to apply the power of Gemini in new compute and latency-sensitive settings.

Rigorous evaluations provided by the Massive Multilingual Text Embedding Benchmark (MMTEB) reveal substantial gains over previous top-performing models advancing the state-of-the-art in performance on multilingual, English, and code evaluations. Beyond strong overall performance, Gemini Embedding particularly excels at classification, clustering and retrieval tasks. The advanced versatile and unified capabilities provided by Gemini Embedding and the ability to precompute representations enables the power of Gemini to be leveraged more broadly by both researchers and developers alike.

\bibliography{main}

\begin{thebibliography}{41}
\providecommand{\natexlab}[1]{#1}
\providecommand{\url}[1]{\texttt{#1}}
\expandafter\ifx\csname urlstyle\endcsname\relax
  \providecommand{\doi}[1]{doi: #1}\else
  \providecommand{\doi}{doi: \begingroup \urlstyle{rm}\Url}\fi

\bibitem[Anil et~al.(2023{\natexlab{a}})Anil, Borgeaud, Wu, Alayrac, Yu,
  Soricut, Schalkwyk, Dai, Hauth, et~al.]{team2023gemini}
R.~Anil, S.~Borgeaud, Y.~Wu, J.-B. Alayrac, J.~Yu, R.~Soricut, J.~Schalkwyk,
  A.~M. Dai, A.~Hauth, et~al.
\newblock Gemini: a family of highly capable multimodal models.
\newblock \emph{arXiv preprint arXiv:2312.11805}, 2023{\natexlab{a}}.

\bibitem[Anil et~al.(2023{\natexlab{b}})Anil, Dai, Firat, Johnson, Lepikhin,
  Passos, Shakeri, Taropa, Bailey, Chen, et~al.]{anil2023palm}
R.~Anil, A.~M. Dai, O.~Firat, M.~Johnson, D.~Lepikhin, A.~Passos, S.~Shakeri,
  E.~Taropa, P.~Bailey, Z.~Chen, et~al.
\newblock Palm 2 technical report.
\newblock \emph{arXiv preprint arXiv:2305.10403}, 2023{\natexlab{b}}.

\bibitem[Asai et~al.(2021)Asai, Kasai, Clark, Lee, Choi, and
  Hajishirzi]{asai2021xor}
A.~Asai, J.~Kasai, J.~H. Clark, K.~Lee, E.~Choi, and H.~Hajishirzi.
\newblock Xor qa: Cross-lingual open-retrieval question answering.
\newblock In \emph{Proceedings of the 2021 Conference of the North American
  Chapter of the Association for Computational Linguistics: Human Language
  Technologies}, pages 547--564, 2021.

\bibitem[Bonifacio et~al.(2022)Bonifacio, Abonizio, Fadaee, and
  Nogueira]{bonifacio2022inpars}
L.~Bonifacio, H.~Abonizio, M.~Fadaee, and R.~Nogueira.
\newblock Inpars: Unsupervised dataset generation for information retrieval.
\newblock In \emph{Proceedings of the 45th International ACM SIGIR Conference
  on Research and Development in Information Retrieval}, pages 2387--2392,
  2022.

\bibitem[Brown et~al.(2020)Brown, Mann, Ryder, Subbiah, Kaplan, Dhariwal,
  Neelakantan, Shyam, Sastry, Askell, Agarwal, Herbert{-}Voss, Krueger,
  Henighan, Child, Ramesh, Ziegler, Wu, Winter, Hesse, Chen, Sigler, Litwin,
  Gray, Chess, Clark, Berner, McCandlish, Radford, Sutskever, and
  Amodei]{brown2020gpt}
T.~B. Brown, B.~Mann, N.~Ryder, M.~Subbiah, J.~Kaplan, P.~Dhariwal,
  A.~Neelakantan, P.~Shyam, G.~Sastry, A.~Askell, S.~Agarwal,
  A.~Herbert{-}Voss, G.~Krueger, T.~Henighan, R.~Child, A.~Ramesh, D.~M.
  Ziegler, J.~Wu, C.~Winter, C.~Hesse, M.~Chen, E.~Sigler, M.~Litwin, S.~Gray,
  B.~Chess, J.~Clark, C.~Berner, S.~McCandlish, A.~Radford, I.~Sutskever, and
  D.~Amodei.
\newblock Language models are few-shot learners.
\newblock In H.~Larochelle, M.~Ranzato, R.~Hadsell, M.~Balcan, and H.~Lin,
  editors, \emph{Advances in Neural Information Processing Systems 33: Annual
  Conference on Neural Information Processing Systems 2020, NeurIPS 2020,
  December 6-12, 2020, virtual}, 2020.

\bibitem[Cer et~al.(2018)Cer, Yang, Kong, Hua, Limtiaco, John, Constant,
  Guajardo-Cespedes, Yuan, Tar, et~al.]{cer2018universal}
D.~Cer, Y.~Yang, S.-y. Kong, N.~Hua, N.~Limtiaco, R.~S. John, N.~Constant,
  M.~Guajardo-Cespedes, S.~Yuan, C.~Tar, et~al.
\newblock Universal sentence encoder for english.
\newblock In \emph{Proceedings of the 2018 conference on empirical methods in
  natural language processing: system demonstrations}, pages 169--174, 2018.

\bibitem[Cormack et~al.(2009)Cormack, Clarke, and
  Buettcher]{cormack2009reciprocal}
G.~V. Cormack, C.~L. Clarke, and S.~Buettcher.
\newblock Reciprocal rank fusion outperforms condorcet and individual rank
  learning methods.
\newblock In \emph{Proceedings of the 32nd international ACM SIGIR conference
  on Research and development in information retrieval}, pages 758--759, 2009.

\bibitem[Dai et~al.(2022)Dai, Zhao, Ma, Luan, Ni, Lu, Bakalov, Guu, Hall, and
  Chang]{dai2022promptagator}
Z.~Dai, V.~Y. Zhao, J.~Ma, Y.~Luan, J.~Ni, J.~Lu, A.~Bakalov, K.~Guu, K.~B.
  Hall, and M.-W. Chang.
\newblock Promptagator: Few-shot dense retrieval from 8 examples.
\newblock \emph{arXiv preprint arXiv:2209.11755}, 2022.

\bibitem[Devlin et~al.(2019)Devlin, Chang, Lee, and Toutanova]{devlin2019bert}
J.~Devlin, M.~Chang, K.~Lee, and K.~Toutanova.
\newblock {BERT:} pre-training of deep bidirectional transformers for language
  understanding.
\newblock In J.~Burstein, C.~Doran, and T.~Solorio, editors, \emph{Proceedings
  of the 2019 Conference of the North American Chapter of the Association for
  Computational Linguistics: Human Language Technologies, {NAACL-HLT} 2019,
  Minneapolis, MN, USA, June 2-7, 2019, Volume 1 (Long and Short Papers)},
  pages 4171--4186. Association for Computational Linguistics, 2019.

\bibitem[Enevoldsen et~al.(2025)Enevoldsen, Chung, Kerboua, Kardos, Mathur,
  Stap, Gala, Siblini, Krzemi{\'n}ski, Winata, et~al.]{enevoldsen2025mmteb}
K.~Enevoldsen, I.~Chung, I.~Kerboua, M.~Kardos, A.~Mathur, D.~Stap, J.~Gala,
  W.~Siblini, D.~Krzemi{\'n}ski, G.~I. Winata, et~al.
\newblock Mmteb: Massive multilingual text embedding benchmark.
\newblock \emph{arXiv preprint arXiv:2502.13595}, 2025.

\bibitem[Feng et~al.(2022)Feng, Yang, Cer, Arivazhagan, and
  Wang]{feng-etal-2022-language}
F.~Feng, Y.~Yang, D.~Cer, N.~Arivazhagan, and W.~Wang.
\newblock Language-agnostic {BERT} sentence embedding.
\newblock In S.~Muresan, P.~Nakov, and A.~Villavicencio, editors,
  \emph{Proceedings of the 60th Annual Meeting of the Association for
  Computational Linguistics (Volume 1: Long Papers)}, May 2022.
\newblock URL \url{https://aclanthology.org/2022.acl-long.62/}.

\bibitem[Gao et~al.(2021)Gao, Yao, and Chen]{gao2021simcse}
T.~Gao, X.~Yao, and D.~Chen.
\newblock Simcse: Simple contrastive learning of sentence embeddings.
\newblock In \emph{Proceedings of the 2021 Conference on Empirical Methods in
  Natural Language Processing}, pages 6894--6910, 2021.

\bibitem[Izacard et~al.(2022)Izacard, Caron, Hosseini, Riedel, Bojanowski,
  Joulin, and Grave]{izacard2021contriever}
G.~Izacard, M.~Caron, L.~Hosseini, S.~Riedel, P.~Bojanowski, A.~Joulin, and
  E.~Grave.
\newblock Unsupervised dense information retrieval with contrastive learning.
\newblock \emph{Transactions on Machine Learning Research}, 2022.

\bibitem[Izmailov et~al.(2018)Izmailov, Podoprikhin, Garipov, Vetrov, and
  Wilson]{izmailov2018averaging}
P.~Izmailov, D.~Podoprikhin, T.~Garipov, D.~Vetrov, and A.~G. Wilson.
\newblock Averaging weights leads to wider optima and better generalization.
\newblock \emph{arXiv preprint arXiv:1803.05407}, 2018.

\bibitem[Jeronymo et~al.(2023)Jeronymo, Bonifacio, Abonizio, Fadaee, Lotufo,
  Zavrel, and Nogueira]{jeronymo2023inpars}
V.~Jeronymo, L.~Bonifacio, H.~Abonizio, M.~Fadaee, R.~Lotufo, J.~Zavrel, and
  R.~Nogueira.
\newblock Inpars-v2: Large language models as efficient dataset generators for
  information retrieval.
\newblock \emph{arXiv preprint arXiv:2301.01820}, 2023.

\bibitem[Jiang et~al.(2023)Jiang, Sablayrolles, Mensch, Bamford, Chaplot,
  de~las Casas, Bressand, Lengyel, Lample, Saulnier, Lavaud, Lachaux, Stock,
  Scao, Lavril, Wang, Lacroix, and Sayed]{jiang2023mistral7b}
A.~Q. Jiang, A.~Sablayrolles, A.~Mensch, C.~Bamford, D.~S. Chaplot, D.~de~las
  Casas, F.~Bressand, G.~Lengyel, G.~Lample, L.~Saulnier, L.~R. Lavaud, M.-A.
  Lachaux, P.~Stock, T.~L. Scao, T.~Lavril, T.~Wang, T.~Lacroix, and W.~E.
  Sayed.
\newblock Mistral 7b, 2023.
\newblock URL \url{https://arxiv.org/abs/2310.06825}.

\bibitem[Jiang et~al.(2024)Jiang, Meng, Yang, Yavuz, Zhou, and
  Chen]{jiang2024vlm2vec}
Z.~Jiang, R.~Meng, X.~Yang, S.~Yavuz, Y.~Zhou, and W.~Chen.
\newblock Vlm2vec: Training vision-language models for massive multimodal
  embedding tasks.
\newblock \emph{arXiv preprint arXiv:2410.05160}, 2024.

\bibitem[Karpukhin et~al.(2020)Karpukhin, Oğuz, Min, Lewis, Wu, Edunov, Chen,
  and tau Yih]{Karpukhin2020DensePR}
V.~Karpukhin, B.~Oğuz, S.~Min, P.~Lewis, L.~Y. Wu, S.~Edunov, D.~Chen, and
  W.~tau Yih.
\newblock Dense passage retrieval for open-domain question answering.
\newblock \emph{ArXiv}, abs/2004.04906, 2020.

\bibitem[Kusupati et~al.(2022)Kusupati, Bhatt, Rege, Wallingford, Sinha,
  Ramanujan, Howard-Snyder, Chen, Kakade, Jain, et~al.]{kusupati2022matryoshka}
A.~Kusupati, G.~Bhatt, A.~Rege, M.~Wallingford, A.~Sinha, V.~Ramanujan,
  W.~Howard-Snyder, K.~Chen, S.~Kakade, P.~Jain, et~al.
\newblock Matryoshka representation learning.
\newblock \emph{Advances in Neural Information Processing Systems},
  35:\penalty0 30233--30249, 2022.

\bibitem[Le and Mikolov(2014)]{le2014distributed}
Q.~Le and T.~Mikolov.
\newblock Distributed representations of sentences and documents.
\newblock In \emph{International conference on machine learning}, pages
  1188--1196. PMLR, 2014.

\bibitem[Lee et~al.(2025)Lee, Roy, Xu, Raiman, Shoeybi, Catanzaro, and
  Ping]{lee2025nvembedimprovedtechniquestraining}
C.~Lee, R.~Roy, M.~Xu, J.~Raiman, M.~Shoeybi, B.~Catanzaro, and W.~Ping.
\newblock Nv-embed: Improved techniques for training llms as generalist
  embedding models.
\newblock \emph{ArXiv}, 2025.
\newblock URL \url{https://arxiv.org/abs/2405.17428}.

\bibitem[Lee et~al.(2024)Lee, Dai, Ren, Chen, Cer, Cole, Hui, Boratko, Kapadia,
  Ding, Luan, Duddu, Abrego, Shi, Gupta, Kusupati, Jain, Jonnalagadda, Chang,
  and Naim]{lee2024gecko}
J.~Lee, Z.~Dai, X.~Ren, B.~Chen, D.~Cer, J.~R. Cole, K.~Hui, M.~Boratko,
  R.~Kapadia, W.~Ding, Y.~Luan, S.~M.~K. Duddu, G.~H. Abrego, W.~Shi, N.~Gupta,
  A.~Kusupati, P.~Jain, S.~R. Jonnalagadda, M.-W. Chang, and I.~Naim.
\newblock Gecko: Versatile text embeddings distilled from large language
  models.
\newblock \emph{arXiv preprint arXiv:2403.20327}, 2024.

\bibitem[Li et~al.(2024)Li, Qin, Xiao, Chen, Luo, Shao, Lian, and
  Liu]{li2024makingtextembeddersfewshot}
C.~Li, M.~Qin, S.~Xiao, J.~Chen, K.~Luo, Y.~Shao, D.~Lian, and Z.~Liu.
\newblock Making text embedders few-shot learners.
\newblock \emph{ArXiv}, 2024.
\newblock URL \url{https://arxiv.org/abs/2409.15700}.

\bibitem[Meng et~al.(2024)Meng, Liu, Joty, Xiong, Zhou, and
  Yavuz]{meng2024sfrembedding}
R.~Meng, Y.~Liu, S.~R. Joty, C.~Xiong, Y.~Zhou, and S.~Yavuz.
\newblock Sfrembedding-mistral: enhance text retrieval with transfer learning.
\newblock \emph{Salesforce AI Research Blog}, 3:\penalty0 6, 2024.

\bibitem[Moiseev et~al.(2023)Moiseev, Abrego, Dornbach, Zitouni, Alfonseca, and
  Dong]{moiseev2023samtone}
F.~Moiseev, G.~H. Abrego, P.~Dornbach, I.~Zitouni, E.~Alfonseca, and Z.~Dong.
\newblock Samtone: Improving contrastive loss for dual encoder retrieval models
  with same tower negatives.
\newblock \emph{arXiv preprint arXiv:2306.02516}, 2023.

\bibitem[Muennighoff et~al.(2023)Muennighoff, Tazi, Magne, and
  Reimers]{muennighoff2023mteb}
N.~Muennighoff, N.~Tazi, L.~Magne, and N.~Reimers.
\newblock Mteb: Massive text embedding benchmark.
\newblock In \emph{Proceedings of the 17th Conference of the European Chapter
  of the Association for Computational Linguistics}, pages 2006--2029, 2023.

\bibitem[Neelakantan et~al.(2022)Neelakantan, Xu, Puri, Radford, Han, Tworek,
  Yuan, Tezak, Kim, Hallacy, et~al.]{neelakantan2022text}
A.~Neelakantan, T.~Xu, R.~Puri, A.~Radford, J.~M. Han, J.~Tworek, Q.~Yuan,
  N.~Tezak, J.~W. Kim, C.~Hallacy, et~al.
\newblock Text and code embeddings by contrastive pre-training.
\newblock \emph{arXiv preprint arXiv:2201.10005}, 2022.

\bibitem[Ni et~al.(2021)Ni, Qu, Lu, Dai, 'Abrego, Ma, Zhao, Luan, Hall, Chang,
  and Yang]{Ni2021LargeDE}
J.~Ni, C.~Qu, J.~Lu, Z.~Dai, G.~H. 'Abrego, J.~Ma, V.~Zhao, Y.~Luan, K.~B.
  Hall, M.-W. Chang, and Y.~Yang.
\newblock Large dual encoders are generalizable retrievers.
\newblock In \emph{Conference on Empirical Methods in Natural Language
  Processing}, 2021.

\bibitem[Ni et~al.(2022)Ni, Abrego, Constant, Ma, Hall, Cer, and
  Yang]{ni2022sentence}
J.~Ni, G.~H. Abrego, N.~Constant, J.~Ma, K.~Hall, D.~Cer, and Y.~Yang.
\newblock Sentence-t5: Scalable sentence encoders from pre-trained text-to-text
  models.
\newblock In \emph{Findings of the Association for Computational Linguistics:
  ACL 2022}, pages 1864--1874, 2022.

\bibitem[Nogueira et~al.(2019)Nogueira, Yang, Lin, and
  Cho]{nogueira2019document}
R.~Nogueira, W.~Yang, J.~Lin, and K.~Cho.
\newblock Document expansion by query prediction.
\newblock \emph{arXiv preprint arXiv:1904.08375}, 2019.

\bibitem[Raffel et~al.(2020)Raffel, Shazeer, Roberts, Lee, Narang, Matena,
  Zhou, Li, and Liu]{raffel2020exploring}
C.~Raffel, N.~Shazeer, A.~Roberts, K.~Lee, S.~Narang, M.~Matena, Y.~Zhou,
  W.~Li, and P.~J. Liu.
\newblock Exploring the limits of transfer learning with a unified text-to-text
  transformer.
\newblock \emph{The Journal of Machine Learning Research}, 21\penalty0
  (1):\penalty0 5485--5551, 2020.

\bibitem[Reddi et~al.(2019)Reddi, Kale, Yu, Holtmann-Rice, Chen, and
  Kumar]{reddi2019stochastic}
S.~J. Reddi, S.~Kale, F.~Yu, D.~Holtmann-Rice, J.~Chen, and S.~Kumar.
\newblock Stochastic negative mining for learning with large output spaces.
\newblock In \emph{The 22nd International Conference on Artificial Intelligence
  and Statistics}, pages 1940--1949. PMLR, 2019.

\bibitem[Reimers and Gurevych(2019)]{reimers2019sentence}
N.~Reimers and I.~Gurevych.
\newblock Sentence-bert: Sentence embeddings using siamese bert-networks.
\newblock In \emph{Proceedings of the 2019 Conference on Empirical Methods in
  Natural Language Processing and the 9th International Joint Conference on
  Natural Language Processing (EMNLP-IJCNLP)}, pages 3982--3992, 2019.

\bibitem[Ruder et~al.(2023)Ruder, Clark, Gutkin, Kale, Ma, Nicosia, Rijhwani,
  Riley, Sarr, Wang, et~al.]{ruder2023xtreme}
S.~Ruder, J.~H. Clark, A.~Gutkin, M.~Kale, M.~Ma, M.~Nicosia, S.~Rijhwani,
  P.~Riley, J.-M. Sarr, X.~Wang, et~al.
\newblock Xtreme-up: A user-centric scarce-data benchmark for under-represented
  languages.
\newblock In \emph{Findings of the Association for Computational Linguistics:
  EMNLP 2023}, pages 1856--1884, 2023.

\bibitem[Suganthan et~al.(2025)Suganthan, Moiseev, Yan, Wu, Ni, Han, Zitouni,
  Alfonseca, Wang, and Dong]{suganthan2025adaptingdecoder}
P.~Suganthan, F.~Moiseev, L.~Yan, J.~Wu, J.~Ni, J.~Han, I.~Zitouni,
  E.~Alfonseca, X.~Wang, and Z.~Dong.
\newblock Adapting decoder-based language models for diverse encoder downstream
  tasks, 2025.
\newblock URL \url{https://arxiv.org/abs/2503.02656}.

\bibitem[Team(2024)]{geminiteam2024gemini15unlockingmultimodal}
G.~Team.
\newblock Gemini 1.5: Unlocking multimodal understanding across millions of
  tokens of context, 2024.
\newblock URL \url{https://arxiv.org/abs/2403.05530}.

\bibitem[Thakur et~al.(2024)Thakur, Ni, Hernandez~Abrego, Wieting, Lin, and
  Cer]{thakur-etal-2024-leveraging}
N.~Thakur, J.~Ni, G.~Hernandez~Abrego, J.~Wieting, J.~Lin, and D.~Cer.
\newblock Leveraging {LLM}s for synthesizing training data across many
  languages in multilingual dense retrieval.
\newblock In \emph{Proceedings of the 2024 Conference of the North American
  Chapter of the Association for Computational Linguistics: Human Language
  Technologies (Volume 1: Long Papers)}, June 2024.
\newblock URL \url{https://aclanthology.org/2024.naacl-long.426/}.

\bibitem[Wang et~al.(2022)Wang, Yang, Huang, Jiao, Yang, Jiang, Majumder, and
  Wei]{wang2022text}
L.~Wang, N.~Yang, X.~Huang, B.~Jiao, L.~Yang, D.~Jiang, R.~Majumder, and
  F.~Wei.
\newblock Text embeddings by weakly-supervised contrastive pre-training.
\newblock \emph{arXiv preprint arXiv:2212.03533}, 2022.

\bibitem[Wang et~al.(2023)Wang, Yang, Huang, Yang, Majumder, and
  Wei]{wang2023improving}
L.~Wang, N.~Yang, X.~Huang, L.~Yang, R.~Majumder, and F.~Wei.
\newblock Improving text embeddings with large language models.
\newblock \emph{arXiv preprint arXiv:2401.00368}, 2023.

\bibitem[Wortsman et~al.(2022)Wortsman, Ilharco, Gadre, Roelofs, Gontijo-Lopes,
  Morcos, Namkoong, Farhadi, Carmon, Kornblith, et~al.]{wortsman2022model}
M.~Wortsman, G.~Ilharco, S.~Y. Gadre, R.~Roelofs, R.~Gontijo-Lopes, A.~S.
  Morcos, H.~Namkoong, A.~Farhadi, Y.~Carmon, S.~Kornblith, et~al.
\newblock Model soups: averaging weights of multiple fine-tuned models improves
  accuracy without increasing inference time.
\newblock In \emph{International conference on machine learning}, pages
  23965--23998. PMLR, 2022.

\bibitem[Zhang et~al.(2023)Zhang, Thakur, Ogundepo, Kamalloo, Alfonso-Hermelo,
  Li, Liu, Rezagholizadeh, and Lin]{zhang2023miracl}
X.~Zhang, N.~Thakur, O.~Ogundepo, E.~Kamalloo, D.~Alfonso-Hermelo, X.~Li,
  Q.~Liu, M.~Rezagholizadeh, and J.~Lin.
\newblock Miracl: A multilingual retrieval dataset covering 18 diverse
  languages.
\newblock \emph{Transactions of the Association for Computational Linguistics},
  11:\penalty0 1114--1131, 2023.

\end{thebibliography}

\clearpage
\section{Full Results}
\begin{table}[ht]
\centering
\caption{Full results of Gemini Embedding on MTEB(Multilingual).}
\resizebox{0.3365\columnwidth}{!}{%
\begin{tabular}{lc}
\toprule
Task Name & Performance \\
\midrule
AILAStatutes & 48.77 \\
AfriSentiClassification & 53.56 \\
AlloProfClusteringS2S.v2 & 56.36 \\
AlloprofReranking & 81.77 \\
AmazonCounterfactualClassification & 88.20 \\
ArXivHierarchicalClusteringP2P & 64.92 \\
ArXivHierarchicalClusteringS2S & 63.84 \\
ArguAna & 86.44 \\
ArmenianParaphrasePC & 96.89 \\
BUCC.v2 & 98.99 \\
BelebeleRetrieval & 90.73 \\
BibleNLPBitextMining & 20.72 \\
BigPatentClustering.v2 & 38.06 \\
BiorxivClusteringP2P.v2 & 53.86 \\
BornholmBitextMining & 51.69 \\
BrazilianToxicTweetsClassification & 28.02 \\
BulgarianStoreReviewSentimentClassfication & 78.13 \\
CEDRClassification & 57.42 \\
CLSClusteringP2P.v2 & 42.68 \\
CSFDSKMovieReviewSentimentClassification & 49.38 \\
CTKFactsNLI & 87.59 \\
CataloniaTweetClassification & 54.51 \\
Core17InstructionRetrieval & 7.69 \\
CovidRetrieval & 79.13 \\
CyrillicTurkicLangClassification & 95.30 \\
CzechProductReviewSentimentClassification & 68.16 \\
DBpediaClassification & 94.76 \\
DalajClassification & 50.47 \\
DiaBlaBitextMining & 87.23 \\
EstonianValenceClassification & 53.52 \\
FaroeseSTS & 86.12 \\
FilipinoShopeeReviewsClassification & 48.45 \\
FinParaSTS & 28.60 \\
FinancialPhrasebankClassification & 88.64 \\
FloresBitextMining & 83.71 \\
GermanSTSBenchmark & 88.09 \\
GreekLegalCodeClassification & 43.76 \\
GujaratiNewsClassification & 92.05 \\
HALClusteringS2S.v2 & 32.00 \\
HagridRetrieval & 99.31 \\
IN22GenBitextMining & 93.75 \\
IndicCrosslingualSTS & 62.87 \\
IndicGenBenchFloresBitextMining & 96.77 \\
IndicLangClassification & 87.69 \\
IndonesianIdClickbaitClassification & 67.00 \\
IsiZuluNewsClassification & 40.53 \\
ItaCaseholdClassification & 73.30 \\
JSICK & 84.99 \\
KorHateSpeechMLClassification & 17.69 \\
KorSarcasmClassification & 60.51 \\
KurdishSentimentClassification & 86.39 \\
LEMBPasskeyRetrieval & 38.50 \\
LegalBenchCorporateLobbying & 95.98 \\
MIRACLRetrievalHardNegatives & 70.42 \\
MLQARetrieval & 84.16 \\
MacedonianTweetSentimentClassification & 71.83 \\
MalteseNewsClassification & 37.38 \\
MasakhaNEWSClassification & 83.55 \\
MasakhaNEWSClusteringS2S & 57.45 \\
MassiveIntentClassification & 81.92 \\
MedrxivClusteringP2P.v2 & 47.16 \\
MultiEURLEXMultilabelClassification & 5.28 \\
MultiHateClassification & 72.47 \\
NTREXBitextMining & 93.64 \\
NepaliNewsClassification & 98.14 \\
News21InstructionRetrieval & 10.26 \\
\bottomrule
\end{tabular}}
\centering
\resizebox{0.33\columnwidth}{!}{%
\begin{tabular}{lc}
\toprule
Task Name & Performance \\
\midrule
NollySentiBitextMining & 68.71 \\
NordicLangClassification & 85.97 \\
NorwegianCourtsBitextMining & 93.42 \\
NusaParagraphEmotionClassification & 56.38 \\
NusaTranslationBitextMining & 77.52 \\
NusaX-senti & 80.31 \\
NusaXBitextMining & 82.52 \\
OdiaNewsClassification & 91.84 \\
OpusparcusPC & 96.62 \\
PAC & 71.68 \\
PawsXPairClassification & 59.99 \\
PlscClusteringP2P.v2 & 74.31 \\
PoemSentimentClassification & 59.66 \\
PolEmo2.0-OUT & 77.53 \\
PpcPC & 95.50 \\
PunjabiNewsClassification & 82.61 \\
RTE3 & 89.55 \\
Robust04InstructionRetrieval & -2.41 \\
RomaniBibleClustering & 43.22 \\
RuBQReranking & 73.84 \\
SCIDOCS & 25.15 \\
SIB200ClusteringS2S & 41.74 \\
SICK-R & 82.75 \\
SNLHierarchicalClusteringP2P & 61.41 \\
STS12 & 81.55 \\
STS13 & 89.89 \\
STS14 & 85.41 \\
STS15 & 90.44 \\
STS17 & 88.58 \\
STS22.v2 & 71.69 \\
STSB & 85.50 \\
STSBenchmark & 89.08 \\
STSES & 81.75 \\
ScalaClassification & 51.85 \\
SemRel24STS & 73.14 \\
SentimentAnalysisHindi & 76.06 \\
SinhalaNewsClassification & 82.29 \\
SiswatiNewsClassification & 62.38 \\
SlovakMovieReviewSentimentClassification & 90.35 \\
SpartQA & 10.30 \\
SprintDuplicateQuestions & 96.90 \\
StackExchangeClustering.v2 & 92.07 \\
StackOverflowQA & 96.71 \\
StatcanDialogueDatasetRetrieRetrieval & 51.11 \\
SwahiliNewsClassification & 66.05 \\
SwednClusteringP2P & 45.84 \\
SwissJudgementClassification & 57.86 \\
T2Reranking & 67.95 \\
TERRa & 63.92 \\
TRECCOVID & 86.32 \\
Tatoeba & 81.97 \\
TempReasonL1 & 2.96 \\
ToxicConversationsClassification & 88.75 \\
TswanaNewsClassification & 53.37 \\
TweetTopicSingleClassification & 71.11 \\
TwitterHjerneRetrieval & 98.02 \\
TwitterURLCorpus & 87.05 \\
VoyageMMarcoReranking & 66.73 \\
WebLINXCandidatesReranking & 10.97 \\
WikiCitiesClustering & 91.63 \\
WikiClusteringP2P.v2 & 28.23 \\
WikipediaRerankingMultilingual & 92.24 \\
WikipediaRetrievalMultilingual & 94.20 \\
WinoGrande & 60.52 \\
XNLI & 85.26 \\
indonli & 60.69 \\
\bottomrule
\end{tabular}
}
\end{table}

\begin{table}[ht]
\centering
\caption{Full results of Gemini Embedding on MTEB(Eng, v2) (left) and MTEB(Code) (right).}
\resizebox{0.3365\columnwidth}{!}{%
\begin{tabular}{lc}
\toprule
Task Name & Performance \\
\midrule
AmazonCounterfactualClassification & 92.69 \\
ArXivHierarchicalClusteringP2P & 64.92 \\
ArXivHierarchicalClusteringS2S & 63.84 \\
ArguAna & 86.44 \\
AskUbuntuDupQuestions & 64.24 \\
BIOSSES & 88.97 \\
Banking77Classification & 94.27 \\
BiorxivClusteringP2P.v2 & 53.86 \\
CQADupstackGamingRetrieval & 70.68 \\
CQADupstackUnixRetrieval & 53.69 \\
ClimateFEVERHardNegatives & 31.06 \\
FEVERHardNegatives & 88.98 \\
FiQA2018 & 61.78 \\
HotpotQAHardNegatives & 87.01 \\
ImdbClassification & 94.98 \\
MTOPDomainClassification & 99.27 \\
MassiveIntentClassification & 88.46 \\
MassiveScenarioClassification & 92.08 \\
MedrxivClusteringP2P.v2 & 47.16 \\
MedrxivClusteringS2S.v2 & 45.01 \\
MindSmallReranking & 32.95 \\
SCIDOCS & 24.04 \\
SICK-R & 82.75 \\
STS12 & 81.55 \\
STS13 & 89.89 \\
STS14 & 85.41 \\
STS15 & 90.44 \\
STS17 & 91.61 \\
STS22.v2 & 68.37 \\
STSBenchmark & 89.08 \\
SprintDuplicateQuestions & 96.90 \\
StackExchangeClustering.v2 & 92.07 \\
StackExchangeClusteringP2P.v2 & 50.91 \\
SummEvalSummarization.v2 & 38.28 \\
TRECCOVID & 86.32 \\
Touche2020Retrieval.v3 & 52.39 \\
ToxicConversationsClassification & 88.75 \\
TweetSentimentExtractionClassification & 69.88 \\
TwentyNewsgroupsClustering.v2 & 57.37 \\
TwitterSemEval2015 & 79.17 \\
TwitterURLCorpus & 87.05 \\
\bottomrule
\end{tabular}}
\quad
\centering
\resizebox{0.33\columnwidth}{!}{%
\begin{tabular}{lc}
\toprule
Task Name & Performance \\
\midrule
AppsRetrieval & 93.75 \\
COIRCodeSearchNetRetrieval & 81.06 \\
CodeEditSearchRetrieval & 81.61 \\
CodeFeedbackMT & 56.28 \\
CodeFeedbackST & 85.33 \\
CodeSearchNetCCRetrieval & 84.69 \\
CodeSearchNetRetrieval & 91.33 \\
CodeTransOceanContest & 89.53 \\
CodeTransOceanDL & 31.47 \\
CosQA & 50.24 \\
StackOverflowQA & 95.92 \\
SyntheticText2SQL & 69.96 \\
\bottomrule
\end{tabular}
}
\end{table}

\begin{table}[ht]
\centering
\caption{Full results of Gemini Embedding on XOR-Retrieve (left) and XTREME-UP (right).}
\resizebox{0.25\columnwidth}{!}{%
\begin{tabular}{lc}
\toprule
Language & Performance \\
\midrule
ar & 91.26 \\
bn & 94.08 \\
fi & 89.17 \\
ja & 86.31 \\
ko & 89.82 \\
ru & 88.61 \\
te & 93.70 \\
\bottomrule
\end{tabular}}
\quad\quad\quad
\centering
\resizebox{0.2\columnwidth}{!}{%
\begin{tabular}{lc}
\toprule
Language & Performance \\
\midrule
as & 69.25 \\
bho & 66.38 \\
brx & 25.66 \\
gbm & 64.87 \\
gom & 65.54 \\
gu & 70.26 \\
hi & 69.06 \\
hne & 68.33 \\
kn & 69.54 \\
mai & 68.39 \\
ml & 70.82 \\
mni & 44.44 \\
mr & 68.77 \\
mwr & 66.49 \\
or & 65.77 \\
pa & 69.55 \\
ps & 61.90 \\
sa & 68.09 \\
ta & 68.57 \\
ur & 64.85 \\
\bottomrule
\end{tabular}
}
\end{table}

\clearpage
\twocolumn[  
    \begin{@twocolumnfalse}
        \section{Contributions and Acknowledgments}
     \end{@twocolumnfalse}
]

\noindent\textbf{Core Contributors} ($^*$: equal contributions)\\
Jinhyuk Lee$^*$\\
Feiyang Chen$^*$\\
Sahil Dua$^*$\\
Daniel Cer$^*$\\
Madhuri Shanbhogue$^*$\\
Iftekhar Naim\\
Gustavo Hern{\'{a}}ndez {\'{A}}brego\\
Zhe Li\\
Kaifeng Chen\\
Henrique Schechter Vera\\
Xiaoqi Ren\\
Shanfeng Zhang\\
Daniel Salz\\
Michael Boratko\\
Jay Han\\
Blair Chen\\
Shuo Huang\\
Vikram Rao\\

\noindent\textbf{Contributors}\\
Paul Suganthan\\
Feng Han\\
Andreas Doumanoglou\\
Nithi Gupta\\
Fedor Moiseev\\
Cathy Yip\\
Aashi Jain\\
Simon Baumgartner\\
Shahrokh Shahi\\
Frank Palma Gomez\\
Sandeep Mariserla\\
Min Choi\\
Parashar Shah\\
Sonam Goenka\\
Ke Chen\\
Ye Xia\\
Koert Chen\\
Sai Meher Karthik Duddu\\
Yichang Chen\\
Trevor Walker\\
Wenlei Zhou\\
Rakesh Ghiya \\

\newpage

\noindent\textbf{Leadership}\\
Zach Gleicher\\
Karan Gill\\
Zhe Dong\\
Mojtaba Seyedhosseini\\
Yunhsuan Sung\\
Raphael Hoffmann\\
Tom Duerig

\clearpage
\onecolumn
\noindent \textbf{Acknowledgement}\\
Anthony Chen, Slav Petrov, Ben Hora, Andrew McCallum, Manzil Zaheer, Lakshman Yagati, Fernando Pereira, Tania Bedrax-Weiss, Nicholas Monath, Enrique Alfonseca, Xinyang Yi, Lichan Hong, Andrew Lee, Lisa Patel, Ayla Karmali, Aditya Kusupati, Andrew Forbes, Scott Crowell, Srini Narayanan, Sean Nakamoto, Roopal Garg, Golnaz Farhadi, Ye Tian, Hongxiang Gu, Huijie Feng, Jiameng Fan, Pelin Dogan Schönberger, Grzegorz Makosa, Mário Lipovský, Peter Ralbovsky, István Gyürki, Yi-Ting Chen, Zhongli Ding, Tanmaya Dabral, Ariel Fuxman, Chun-Ta Lu, Stein Xudong Lin, Yi Luan, Howard Zhou, Michael Kwong, Ting Liu

\end{document}